\documentclass[11pt]{article}

\usepackage{amssymb,amsmath}
\usepackage{color}
\usepackage{graphicx}
  \DeclareGraphicsExtensions{.jpeg,.png,.jpg,.eps}
\usepackage{tikz}
\usepackage{epstopdf}

\def\c{{\bf c}}

\def\i{{\bf i}}

\def\y{{\bf y}}

\def\A{{\cal A}}
\def\B{{\cal B}}
\def\C{{\cal C}}

\def\F{{\cal F}}

\def\H{{\cal H}}

\def\M{{\cal M}}
\def\N{{\cal N}}
\def\P{{\cal P}}

\def\S{{\cal S}}

\def\R{{\mathbb R}}

\def\al{\alpha}
\def\d{\delta}
\def\D{\Delta}
\def\e{\epsilon}

\def\l{\lambda}

\def\om{\omega}
\def\OM{\Omega}
\def\r{\rho}
\def\s{\sigma}
\def\SI{\Sigma}
\def\t{\tau}
\def\th{\theta}

\def\boldeta{{\boldsymbol \eta}}

\def\mai{m \ap \infty}

\def\ap{\rightarrow}

\def\seq{\subseteq}

\def\bi{\{0,1\}}

\def\bp{\{-1,1\}}

\def\imp{\; \Longrightarrow \;}

\def\fa{\; \forall}

\def\st{\mbox{ s.t. }}
\def\sg{\mbox{sign}}
\def\nm{\Vert}

\renewcommand{\iff}{\mbox{$\; \; \Longleftrightarrow \; \;$}}
\renewcommand{\and}{\mbox{$\wedge$}}

\newcommand{\bc}{\begin{center}}
\newcommand{\ec}{\end{center}}
\newcommand{\be}{\begin{equation}}
\newcommand{\ee}{\end{equation}}
\newcommand{\bd}{\begin{displaymath}}
\newcommand{\ed}{\end{displaymath}}
\newcommand{\ba}{\begin{array}}
\newcommand{\ea}{\end{array}}
\newcommand{\ben}{\begin{enumerate}}
\newcommand{\een}{\end{enumerate}}
\newcommand{\bit}{\begin{itemize}}
\newcommand{\eit}{\end{itemize}}
\newcommand{\beq}{\begin{eqnarray}}
\newcommand{\eeq}{\end{eqnarray}}
\newcommand{\btab}{\begin{tabular}}
\newcommand{\etab}{\end{tabular}}
\newcommand{\bfig}{\begin{figure}}
\newcommand{\efig}{\end{figure}}
\newcommand{\btp}{\begin{tikzpicture}}
\newcommand{\etp}{\end{tikzpicture}}

\newcommand{\argmin}{\operatornamewithlimits{argmin}}

\newcommand{\nmm}[1]{ \nm #1 \nm }
\newcommand{\nmeu}[1]{ \nm #1 \nm_2 }
\newcommand{\nmeusq}[1]{ \nm #1 \nm_2^2 }

\newcommand{\IP}[2]{ \langle #1 , #2 \rangle }

\newcommand{\halmos}{\hfill $\Box$}

\newcommand{\supp}{\mbox{supp}}

\def\xh{\hat{x}}

\def\nmsl1{\nm_{{\rm SL1}}}

\definecolor{verm}{rgb}{0.6,0.2,0.2}
\definecolor{purp}{rgb}{0.3,0.1,0.6}
\definecolor{purple}{rgb}{0.4,0.0,0.6}
\definecolor{bggreen}{rgb}{0.1,0.3,0.1}
\definecolor{dgreen}{rgb}{0.1,0.6,0.1}
\definecolor{black}{rgb}{0.0,0.0,0.0}
\definecolor{crim}{rgb}{0.3,0.1,0.1}
\definecolor{dred}{rgb}{0.5,0.1,0.1}

{\bf}{\it}
\newtheorem{definition}{Definition}{\bf}{\it}
{\bf}{\rm}
\newtheorem{lemma}{Lemma}{\bf}{\it}
\newtheorem{theorem}{Theorem}{\bf}{\it}
{\bf}{\it}
{\bf}{\it}
{\bf}{\rm}

\def\supp{{\rm supp}}
\def\xh{\hat{x}}

\def\Jh{\hat{J}}

\def\vcd{\mbox{VC-dim}}

\begin{document}

\title{
An Approach to One-Bit Compressed Sensing\\
Based on Probably Approximately Correct Learning Theory
}

\author{
Mehmet Eren Ahsen and Mathukumalli Vidyasagar
\thanks{
The research of MV was supported by the National Science Foundation
under Award \# ECCS-1306630 and by the Department of Science and
Technology, Government of India.
}
}

%

\maketitle

\begin{abstract}
In this paper, the problem of one-bit compressed sensing (OBCS) is
formulated as a problem in probably approximately correct (PAC) learning.
It is shown that the Vapnik-Chervonenkis (VC-) dimension
of the set of half-spaces in $\R^n$ generated by $k$-sparse vectors
is bounded below by $k \lg (n/k)$ and above by $2k \lg (n/k)$, plus
some round-off terms.
By coupling this estimate with well-established results in PAC
learning theory, we show that a consistent algorithm can recover
a $k$-sparse vector with $O(k \lg (n/k))$ measurements,
given only the signs of the measurement vector.
This result holds for \textit{all} probability measures on $\R^n$.
It is further shown that random sign-flipping errors result only in an increase
in the constant in the $O(k \lg (n/k))$ estimate.
Because constructing a consistent algorithm is not straight-forward,
we present a heuristic based on the $\ell_1$-norm support vector
machine, and illustrate that its computational performance is superior
to a currently popular method.
\end{abstract}

\section{Introduction}\label{sec:intro}

The field of ``compressed sensing'' has become very popular in recent years,
with an explosion in the number of papers.
Stated briefly, the core problem in compressed sensing is to recover a
high-dimensional sparse (or nearly sparse) vector $x$ from a small number
of measurements of $x$.
In the traditional problem formulation, the measurements are \textit{linear},
consisting of $m$ real numbers
$y_i = \IP{ a_i }{ x }, i = 1 , \ldots , m$,
where the measurement vectors $a_i \in \R^n$ are chosen by the learner.
More recently, attention has focused on so-called one-bit compressed sensing,
referred to hereafter as OBCS in the interest of brevity.
In OBCS the measurements consist, not of the inner products
$\IP{ a_i }{ x }$, but rather \textit{just the signs of these
inner products}, i.e., a ``one-bit'' representation of these measurements.
In much of the OBCS literature,
the vectors $a_i$ are chosen at random from some specified
probability distribution, often Gaussian.
Because the sign of $\IP{a_i}{x}$ is unchanged if $x$ is replaced
by any positive multiple of $x$, it is obvious that under this model,
one can at best 
aspire to recover the unknown vector $x$ only to within a positive
multiple, or equivalently, to recover the normalized vector
$x/\nmeu{x}$.
This limitation can be overcome by choosing
the measurements to consist of the signs of inner
products $\IP{ a_i }{ x } + b_i$, where again $a_i,b_i$
are selected at random.

The current status of OBCS is that while
several algorithms have been proposed, theoretical analysis
is available for only a few algorithms.
Moreover, in cases where theoretical analysis is available, the sample
complexity is extremely high.
In the present paper, we interpret OBCS as a problem in
probably approximately correct (PAC) learning theory, which is
a well-established branch of statistical learning theory.
By doing so, we are able to draw from the wealth of results that
are already available, and thereby address many of the currently
outstanding issues.
In PAC learning theory, a central role is played by the so-called
Vapnik-Chervonenkis, or VC-dimension of the collection of concepts
to be learned.
The principal result of the present paper is
that the VC-dimension
of the set of half-planes in $\R^n$ generated by $k$-sparse vectors
is bounded below by $k \lg (n/k)$ and above by $2k \lg (n/k)$,
plus some roundoff terms.
Using this bound, we are able to
establish the following results for the case where $x \in \R^n$
has no more than $k$ nonzero components:
\bit
\item In principle, OBCS is possible whenever
the measurement vector $(a_i,b_i)$ is drawn at random from
\textit{any arbitrary\/} probability distribution.
Moreover, if a consistent algorithm\footnote{This term is standard 
in statistical learning theory and is defined later.}
can be devised, then the number of measurements is $O(k \ln (n/k))$.
\item There is also a \textit{lower bound} on the OBCS problem.
Specifically, there exists a probability distribution on $\R^n$ such that, if
$(a_i,b_i)$ is drawn at random from this distribution, then
the number of measurements required to learn each
$k$-sparse $n$-dimensional vector is bounded \textit{below} by
$\OM(k \ln (n/k))$.
\item In view of the intractability of constructing a consistent
algorithm for this problem,
an algorithm based on the $\ell_1$-norm support vector machine 
is proposed for recovering the unknown sparse vector $x$ approximately.
The algorithm is evaluated on a test problem and it is shown that it performs
better than a currently popular method.
\item It is shown that
PAC learning with finite VC-dimension is robust under random
flipping of labels, even when the flipping probability is not known.
Thus, OBCS is still possible in the case where
the actual information available to the learner consists of
the sign of $\IP{ a_i }{ x }$ \textit{which is then passed through
a binary symmetric channel} that flips 0 to 1 and vice versa with
some probability $\al < 0.5$.
Moreover, the number of measurements required is still $O(k \ln (n/k))$,
but with a larger constant under the $O$ symbol.
\item
If the samples $a_i$ are not independent, but
are $\beta$-mixing, learning is still possible, and explicit estimates
are available for the rate of learning.
\eit

The paper is organized as follows:
In Section \ref{sec:OBCS}, a brief review is given of some recent
papers in OBCS.
In Section \ref{sec:prelim}, some parts of PAC learning theory that
are relevant to OBCS are reviewed.
In particular, it is shown how OBCS can be formulated as a problem
in PAC learning, so that OBCS can be addressed by finding upper bounds
on the VC-dimension of half-spaces generated by $k$-sparse vectors.
In Section \ref{sec:VC}, both upper and lower bounds are derived 
for the VC-dimension of half-spaces generated by $k$-sparse vectors.
In Section \ref{sec:noisy}, the standard results in PAC learning
theory, namely that concept classes with finite VC-dimension are PAC
learnable, are extended to the case where measurements are noisy.
While some such results are known in the literature, they require
the probability of mislabelling to be known; no such assumption is made here.
In Section \ref{sec:alg}, we first present a conceptual algorithm
arising from applying PAC learning theory to OBCS.
However, since this algorithm is not computationally feasible,
we then present a tractable algorithm based on the $\ell_1$-norm
support vector machine.
A numerical example is presented in Section \ref{sec:exam}, where it is
shown that our suggested algorithm performs better than 
a currently popular method.
Finally, Section \ref{sec:disc} contains a discussion of
some issues that merit further investigation.

\section{Brief Review of One-Bit Compressed Sensing}\label{sec:OBCS}

By now there is a substantial literature regarding the traditional
compressed sensing formulation, out of which
only a few references are cited here in the interests of brevity.
Book-length treatments of compressed sensing can be found in
\cite{Elad10,FR13,Rish-Grabarnik15,HTW15}.
Amongst these, \cite{FR13} contains a thorough discussion of virtually
aspect of compressed sensing theory.
A recent volume \cite{Eldar-Kutyniok12} is a compendium of articles
on a variety of topics.
The first paper in this volume \cite{DDEK12} is a survey of the basic
results in compressed sensing.
Another paper \cite{NRWY12} provides a very general framework
for sparse regression that can be used, among other things, to analyze
compressed sensing algorithms.
Each of these papers contains an extensive bibliography.

Throughout this paper, $n$ denotes some fixed and large integer.
For $x \in \R^n$, let $\supp(x)$ denote the support of a vector, 
and let $\SI_k$ denote the set of $k$-sparse vectors in $\R^n$; that is,
\bd
\supp(x) := \{ i : x_i \neq 0 \} ,
\SI_k := \{ x \in \R^n : | \supp(x) | \leq k \} .
\ed
Suppose $x \in \R^n$ is $k$-sparse, that is, $x \in \SI_k$,
where both $n$ and $k$ are known integers with $k \ll n$.
The basic problem in compressed sensing is
to design an $m \times n$ matrix $A$ where $m \ll n$, together with
a decoder map $\D: \R^m \ap \R^n$ such that $\D(Ax) = x$ for all $x \in \SI_k$,
that is, $x$ can be recovered exactly from the $m$-dimensional
vector of linear measurements $y = Ax$.
Variations of the problem include the case where $x$ is only
``nearly sparse,'' and/or $y = Ax + \eta$ where $\eta$ is a measurement noise.
By far the most popular method for recovering a sparse vector is 
$\ell_1$-norm minimization.
If $y = Ax + \eta$, the approach is to define
\be\label{eq:11a}
\D(y) = \xh := \argmin_z \nmm{z}_1 \st \nmeu{y - Az} \leq \e ,
\ee
where $\e$ is a known upper bound on $\nmeu{\eta}$.
In a series of papers by Cand\`{e}s, Tao, Donoho, and others, it is
demonstrated that if the matrix $A$ is chosen so as to satisfy the so-called
Restricted Isometry Property (RIP), then the decoder $\D$ defined in
\eqref{eq:11a} produces a good approximation to $x$, and recovers it
exactly if $x \in \SI_k$ and $\eta = 0$.
See for example \cite{Candes-Tao05,CRT06b,Donoho06a,Donoho06b},
as well as the survey paper \cite{DDEK12} and the comprehensive book
\cite{FR13}.
Moreover, it is shown in \cite{Candes-Tao05} that if 
the elements of $A$ are samples of independent and identically distributed
(i.i.d.) normal random variables (denoted by $a_{ij} \sim \N(0,1)$),
then with probability the resulting normalized matrix
$(1/\sqrt{m}) A$ satisfies the RIP.


The remainder of the section is devoted to a discussion of the
one-bit compressed sensing (OBCS) problem.
One bit compressed sensing is introduced in \cite{Bouf-Bar08}.
In that paper, it is assumed that the measurement $y_i$ equals the bipolar
quantity $y_i = \sg( \IP{ a_i }{ x } )$, as opposed to the real number
$\IP{ a_i }{ x }$.
Because the measurements remain invariant if $x$ is replaced by
any positive multiple of $x$,
there is no loss of generality in assuming that $\nmeu { x } = 1$.
A greedy algorithm called ``renormalized fixed point iteration''
is introduced, as follows:
\bd
\xh := \argmin_z \nm z \nm_1 + \lambda \sum_{i=1}^m f( | y_i (A)_i | )
\st \nmeu { z } = 1 ,
\ed
where the regularizing function $f( \cdot )$ is defined by
\bd
f( \al ) := \left\{ \ba{ll}
\al^2 / 2 & \mbox{if } \al < 0 , \\
0 & \mbox{if } \al \geq 0 . \ea \right.
\ed
The optimization problem is non-convex due to the constraint
$\nmeu { z } = 1$.
Only simulations are provided, but no theoretical results.

In \cite{Gupta-et-al10}, the focus is on recovering the support
set of the unknown vector $x$ from noise-corrupted measurements of
the form $\sg( \IP{ a_i }{ x } + \eta_i )$, where the noise
vector $\boldeta$ consists of pairwise independent Gaussian signals.
A non-adaptive algorithm is presented that makes use of Hoeffding's
inequality applied to the expected value of the covariance of the signs
of two Gaussian random variables.
An adaptive algorithm is also presented.
In \cite{Boufounos09}, a new greedy algorithm is presented called ``matched
signed pursuit.''
The optimization problem is not convex; as a result there are no theoretical
resuts.
The algorithm is similar to the CoSaMP algorithm for the conventional
compressed sensing problem \cite{Needell-Tropp08}.

In \cite{Jacques-et-al13}, one begins with a constant $\e_{{\rm opt}}$
that satisfies
\bd
\e_{{\rm opt}} \geq \frac{ k }{ 2e m + 2 k^{3/2} },
\ed
where $e$ denotes the base of the natural logarithm.
Then the following result is shown:
Let $A \sim \N^{m \times n}(0,1)$ consist of $mn$ 
pairwise independent normal random variables,
and let $y_i = \sg ( \IP{ a_i }{ x } )$.
Fix $\e > 0$ and $\delta \in (0,1)$.
If the number of measurements $m$ satisfies
\bd
m \geq \frac {2}{\e} \left( 2k \ln n + 4k \ln \frac{17}{\e}
+ \ln \frac{1}{ \eta } \right) .
\ed
then for every pair $x,s \in \SI_k$,
\bd
\sg(Ax)=\sg(As)  \implies \nmeu{ x - s } \leq \e,
\ed
with probability $\geq 1 - \eta$.
In words, this result means that if we can find a $k$-sparse vector
$s$ that is consistent with the observation vector $y$, then $s$
is close to $x$. 
In fact, $s$ can be made as close to $x$ as desired by increasing
the number of measurements $m$.
Unfortunately, this result is not practical because finding such a vector $s$
is equivalent to finding a minimal $\ell_0$-norm solution consistent
with the observations, which is known to be an NP-hard problem
\cite{Natarajan95}.

In \cite{Plan-Versh13a}, the authors focus on vectors $x \in \R^n$
that satisfy an inequality of the form $\nmm{x}_1 / \nmeu{x} \leq s$.
Note that if $x$ is $s$-sparse, then it satisfies the above inequality,
though of course the converse is not true.
Thus they use the ratio $\nmm{x}_1 / \nmeu{x}$ as a proxy for $\nmm{x}_0$.
They choose measurement vectors $a_i \in \R^n$ according to the Gaussian
distribution, or more generally, any radially invariant distribution;
this means that, under the chosen probability distribution on the
vector $a \in \R^n$, the normalized vector $a/\nmeu{a}$ is uniformly
distributed on the sphere $S^{n-1} \seq \R^n$.
With these randomly generated measurement vectors, the measured
quantities are $y_i = \sg( \IP{a_i}{x} )$.
The authors propose to estimate $x$ via
\be\label{eq:20}
\xh := \argmin_z \nm z \nm_1 \st \sg(\IP{a_i}{z}) = y_i \fa i ,
\sum_{i=1}^m | \IP{a_i}{z} | = m ,
\ee
where $m$ is the number of measurementss.
They show that if
\bd
\delta > C \left( \frac{s}{m} \ln (2n/s) \ln (2n/m + 2m/n) \right)^{1/5}
\ed
for some universal constant $C$, then
with probability $\geq 1 - \exp(-c \delta m)$ where $c$ is another universal
constant, it is true that
\be\label{eq:21}
\left \nm { \frac { x }{ \nmeu { x } } - \frac { \xh }{ \nmeu { \xh } } } \right \nm_2\leq \delta 
\ee
for all $x \in \R^n$ such that $\nm x \nm_1 / \nmeu { x } \leq \sqrt{s}$.
Although this is the first proposed convex algorithm to recover $x$,
the number of measurement $m$ is $O(\d^{-5})$.
From \eqref{eq:21} we see that if we are able to carry out the
$\ell_0$-norm minimization, then $m$ is $O(\d^{-1})$.
It is still an open question whether or not a practical algorithm 
can achieve this optimal dependence on $\d$ in \eqref{eq:21}.

In \cite{Ai-et-al14} the theory is extended to non-Gaussian noise signals
that are sub-Gaussian.
In \cite{Plan-Versh13b},
it is assumed that the measurements could be noisy and that
\bd
E (y_i) = \th ( \IP{ a_i }{ x } ) ,
\ed
where $\th : \R \ap [-1 , 1]$ is an unknown function.
If $\th(\al) = \tanh(\al/2)$, then the problem is one of logistic
regression, whereas if $\th(\al) = \sg(\al)$, then the problem becomes OBCS.
A probabilistic approach is proposed, which has the advantage that
the resulting optimization problem is convex.
However, the disadvantage is that the number of measurements $m$
is $O(\d^{-6})$ where $\d$ is the probability that the algorithm may fail.
The large negative exponent of $\d$ makes the algorithm somewhat impractical.

In all of the papers discussed until now, the measurement vector $y_i$
equals $\sg( \IP{a_i}{x})$ for suitably generated random vectors $a_i$.
As mentioned above, with such a set of measurements one can at best
aspire to recover only the normalized unknown vector $x / \nmeu{x}$.
In \cite{KSW16}, it is proposed to overcome this limitation by
changing the \textit{linear} measurements to \textit{affine} measurements.
Specifically, the measurements in \cite{KSW16} are of the form
$y_i = \sg( \IP{a_i}{x} + b_i )$, where
$a_{ij} \sim \N(0,1)$,\footnote{Recall that each $a_i$ is an $n$-vector.}
and $b_i \sim \N(0,\t^2)$ where $\t$ is some specified constant.
If a prior upper bound $R$ for $\nmeu{x}$ is available, then it is
possible to choose $\t = R$.
Then the optimization problem in \eqref{eq:20} is modified to\footnote{Note
that $v$ here equals $u/\t$ in \cite[Eq.\ (6)]{KSW16}.
Also, since $\d$ is used to denote the confidence of a PAC learning algorithm
later in the paper, we use $\al$ instead of $\d$ as in the cited equation.}
\be\label{eq:22}
( \xh, \hat{v} ) := \argmin_{z, v} [ \nm {z} \nm_1 + \t |v| ]
\st \sg( \IP{a_i}{z} + b_i v ) = y_i \fa i ,
\sum_{i=1}^m | \IP{a_i}{z} + b_i v | = m .
\ee
It is evident that the above formulation is similar to the formulation
in \cite{Plan-Versh13a} applied to the augmented vector $(x,v) \in \R^{n+1}$.
The following result is shown in \cite[Theorem 4]{KSW16}:
Fix $\t, R, \al$ such that $\al < \min \{ 1 , \t/2 \}$.
If
\bd
m \geq C \left( \frac{ \sqrt{R^2 + \t^2} }{\al} \right)^5
\log^2 \left( \frac{2n}{x} \right) ,
\ed
then for all vectors $x \in \R^n$ with $\nmm{x}_1 / \nmeu{s} \leq \sqrt{s}$,
the solution $(\xh , \hat{v})$ to the optimization problem in \eqref{eq:22}
satisfies
\bd
\nmeu{ (\xh/ \hat{v}) - x } \leq \frac{ 4 \sqrt{R^2 + \t^2} } {\t} \al ,
\ed
with a probability exceeding 
\bd
1 - C \exp \left( - \frac{ c \al m}{ \sqrt{R^2 + \t^2} } \right) ,
\ed
where $C$ and $c$ are universal constants.
If $R$ is a known prior upper bound for $\nmeu{x}$, then one can choose
$\t = R$ in the above, in which the bound simplifies to
\bd
\nmeu{ (\xh/ \hat{v}) - x } \leq 4 \sqrt{2} \al .
\ed


\section{Preliminaries}\label{sec:prelim}

In this section we present some preliminary results, while the
main results are presented in the next section.
As shown below, the one-bit compressed sensing (OBCS) problem can
be naturally formulated as a problem in probably approximately correct (PAC)
learning.
In fact, several of the approaches proposed thus far for solving the OBCS
problem are similar to existing methods in PAC learning, but do not
take full advantage of the power and generality of PAC learning theory.
Some of the things that ``come for free'' in PAC learning theory are:
explicit estimates for the number of measurements $m$, ready
extension to the case where successive measurement vectors $a_i$ are
not independent but form a $\beta$-mixing process, and ready extension
to the case of noisy measurements.
However, the PAC learning approach does not readily lend itself
to the formulation of \textit{efficiently computable} algorithms.
This issue is addressed in Section \ref{sec:alg}.

\subsection{Brief Introduction to the PAC Learning Problem}\label{ssec:PAC}

In this subsection, we give a brief introduction to PAC learning theory.
Probably approximately correct (PAC) learning theory can be said to
have originated with the paper \cite{Valiant84}.
%
By now the fundamentals of PAC learning theory are well-developed,
and several book-length treatments
are available, including \cite{Vapnik98,AB99,MV-97,MV-03}.
The theory encompasses a wide variety of learning situations.
However, OBCS is aligned closely with the most basic
version of PAC learning, known as concept learning,
which is formally described next.

The concept learning problem formulation includes the following 
``ingredients'':
\bit
\item An underlying set $X$.
\item A $\s$-algebra $\S$ of subsets of $X$.
\item A collection $\C \seq \S$, known as the ``concept class.'' 
\item A family of probability measures $\P$ on $X$.
\eit
Usually $X$ is a metric space and $\S$ is the Borel $\s$-algebra on $X$.
The family or probability measures $\P$ can range from a single set
$\{ P \}$, to $\P^*$, the set of \textit{all} probability measures
on the set $X$.
If $\P$ is a singleton set $\{P\}$, then the problem is known as
``fixed distribution'' learning, whereas if $\P = \P^*$,
then the problem is known as ``distribution-free'' learning.

Learning takes place as follows:
A fixed but unknown set $T \in \C$, known as the ``target concept,'' is chosen.
If $\P$ consists of more than one probability measure, a fixed but unknown
probability measure $P \in \P$ is also chosen.
Then random samples $\{ c_1 , c_2 , \ldots \}$ are generated independently
in accordance with the chosen distribution $P$.
This is
the basic version of PAC learning studied in \cite{Vapnik98,AB99,MV-97}.
The case where the sample sequence can exhibit dependence, for example
if they come from a Markov process with the stationary distribution $P$,
is studied in \cite{MV-03}.
Using the sample $c_i$, an ``oracle'' generates a ``label'' $y_i \in \bi$.
In the case of noise-free measurements,
$y_i = I_T(c_i)$,
where $T$ is the fixed but unknown target concept, and $I_T(\cdot)$
denotes the indicator function of the set $T$.
Thus the oracle informs the learner whether or not the training sample
$c_i$ belongs to the unknown target concept $T$.
After $m$ such samples are drawn and labelled, the learner makes
use of the set of labelled samples
$\{ (c_i, y_i) \}_{i=1}^m \in ( X \times \bi )^m$ to generate
a ``hypothesis,'' or an approximation to the unknown target concept $T$.
The case where the label $y_i$ is a noisy version of $I_T(c_i)$ is studied
in Section \ref{sec:noisy}.

In statistical learning theory,
an ``algorithm'' is any indexed collection of maps $\{ A_m \}_{m \geq 1}$
where
\bd
A_m : \left( X \times \bi \right)^m \ap \C .
\ed
In other words, an algorithm is any systematic procedure for taking a 
finite sequence of
labelled samples, and returning an element of the concept class $\C$.
The issue of whether $A_m$ is efficiently computable is ignored
in statistical learning theory.
The concept
\bd
G_m(\y;\c) := A_m \left( \{ ( c_i , y_i) \}_{i = 1}^m \right)
\ed
is called the ``hypothesis'' generated by the first $m$ samples
when the sample sequence is $\c = (c_1 , \ldots , c_m)$, and
the label sequence is $\y = (y_1 , \ldots , y_m)$.
In the interests of reducing clutter, we will use $G_m$ in the place of
$G_m(T;\c)$ unless the full form is needed for clarity.
Note that $A_m$ is a deterministic map, but $G_m$ is random because
it depends on the random learning sequence $\{ c_i \}$.

To measure how well the hypothesis $G_m$ approximates the unknown
target concept $T$, we use the \textbf{generalization error} defined by
\be\label{eq:gen-error}
J(T,G_m) = E[ | I_T(x) - I_{G_m}(x) | , P ] .
\ee
Thus $J(T,G_m)$ is the expected value of the difference between the
indicator function $I_T(\cdot)$ and the label generated by the oracle
with the input $I_{G_m}(\cdot)$.
Note that both $I_T$ and $I_{G_m}$ assume values in $\bi$.
Hence $J(T,G_m)$ is also the probability that, when a random test sample
$x \in X$ is generated in accordance with the probability distribution $P$,
the sample is \textit{misclassified} by the hypothesis $G_m$, in the sense that
$I_T(x) \neq I_{G_m}(x)$.

The key quantity in PAC learning theory is the \textbf{learning rate},
defined by
\be\label{eq:rate}
r(m,\e) := \sup_{P \in \P}
\sup_{T \in \C} P^m \{ \c \in X^m : J ( T , G_m ) > \e \} .
\ee
Therefore $r(m,\e)$ is the worst-case measure, over all probability 
distributions in $\P$ and all target concepts in $\C$,
of the set of ``bad'' samples $\c = (c_1 , \ldots , c_m)$
for which the corresponding hypothesis $G_m$ has a generalization
error larger than a prespecifie threshold $\e$.
Thus, after $m$ samples are generated together with their labels,
and the hypothesis $G_m$ is generated using the algorithm, it can
be asserted with confidence $1 - r(m,\e)$ that $G_m$ will correctly
classify the next randomly generated test sample with probability of at
least $1 - \e$.

\begin{definition}\label{def:PAC}
An algorithm $\{ A_m \}$ is said to be {\bf probably approximately
correct (PAC)} if $r(m,\e) \ap 0$ as $\mai$, for every fixed $\e > 0$.
The concept class $\C$ is said to be {\bf PAC learnable} under
the family of probability measures $\P$ is there exists a PAC algorithm.
\end{definition}

The objective of statistical learning theory is to determine conditions
under which there exists a PAC algorithm for a given concept class.

\subsection{OBCS as a Problem in PAC Learning}

In order to embed the problem of one-bit compressed sensing into
the framework of concept learning, we proceed as follows.
We begin with the case where
the measurements are of the form $\sg ( \IP{ a_i }{ x } )$
where the $a_i$ are chosen at random according to some arbitrary probability
distribution, \textit{which need not be the Gaussian.}
Observe now that the closed half-space
\bd
H(x) := \{ z \in \R^n : \IP{z}{x} \geq 0 \} 
\ed
determines the vector $x$ uniquely to within a positive scalar multiple.
Conversely, the vector $x$ uniquely determines the corresponding half-space
$H(x)$, which remains invariant if $x$ is replaced by a positive multiple
of $x$.
Thus the OBCS problem can be posed as that of
determining the half-space $H(x)$ given the measurements
$\sg( \IP{ a_i }{ x } ), i = 1 , \ldots , m$ where the $a_i$
are selected at random in accordance with some probability measure $P$.
Moreover, the one-bit measurement $\sg( \IP{ a_i }{ x } )$
equals $2 I_{H(x)}(a_i) - 1$, where $I_{H(x)} (\cdot )$ denotes the indicator
function of the half-space $H(x)$.
Therefore, to within a simple affine transformation, the OBCS
problem becomes that of determining an unknown half-space
$H(x)$ from labelled samples $(a_i, I_{H(x)}(a_i)) , i = 1 , \ldots , m$,
where the samples $a_i$ are generated at random according to some
prespecified probability measure.
This is a PAC learning problem where the various entities are as follows:
\bit
\item The underlying space $X$would be $\R^n$.
\item The $\s$-algebra $\S$ would be the Borel $\s$-algebra on $\R^n$.
\item The concept class $\C$ would be the collection of all half-spaces
$\{ H_k^n ( x) \}$ where
\be\label{eq:4a}
H_k^n (x) = \{ a \in R^n : \IP{a}{x} \geq 0 \}
\ee
as $x$ varies over $\SI_k$, the set of $k$-sparse vectors in $\R^n$,
\item The family of probability measures $\P$ can either be a singleton
$\{ P \}$ where $P$ is specified \textit{a priori}, or $\P^*$,
the family of \textit{all} probability measures on $\R^n$,
or anything in-between.
\eit

When measurements are of the type $\sg ( \IP{ a_i }{ x } )$,
it is inherently impossible to determine the unknown vector $x$, except
to within a positive scalar multiple.
This is addressed by changing the measurements to be
of the form $\sg ( \IP{ a_i }{ x } + b_i)$,
as suggested in \cite{KSW16}.
Some slight modifications are required to address this modified formulation.
In this case the various entities are as follows:
\bit
\item The underlying space $X$ would equal $\R^{n+1}$.
\item The $\s$-algebra $\S$ would be the Borel $\s$-algebra on $\R^{n+1}$.
\item The concept class $\C$ would be the collection of all half-spaces
$\{ H_k^{n+1} \}$ where
\be\label{eq:4b}
H_k^{n+1} ( x ) = \{ (a , b) \in \R^{n+1} : \IP{a}{x} 
+ b \geq 0 \}
\ee
as $x$ varies over $\SI_k$.
\item The family of probability measures $\P$ can either be a singleton
$\{ P \}$ where $P$ is specified \textit{a priori}, or $\P^*$,
the family of \textit{all} probability measures on $\R^n$,
or anything in-between.
\eit
For a given $x \in \SI_k$, if $a \in \R^n$ belongs to the half-space
$H_k^n(x)$, then the vector $(a,0) \in \R^{n+1}$ belongs to
$H_k^{n+1}(x)$.
However, the half-space $H_k^{n+1}(x)$ can also contain vectors of
the form $(a,b)$ with $b \neq 0$.

\subsection{Interpretation of the Generalization Error}

In the traditional PAC learning problem formulation, the quantity of
interest is the generalization error defined in \eqref{eq:gen-error},
namely
\bd
J(T,G_m) = E[ | I_T(x) - I_{G_m}(x) | , P ] .
\ed
Given two sets $A,B \in \S$, let us define their \textbf{symmetric
difference} by
\bd
A \D B := (A \cup B) \setminus (A \cap B) = (A \cap B^c) \cup (A^c \cap B) ,
\ed
where $A^c$ denotes the complement of $A$.
Thus $A \D B$ consists of the points that belong to precisely one but
not the other set.
Now let us define the quantity
\bd
d_P(A,B) := P(A \D B).
\ed
Then $d_P$ is a pseudometric on $\S$, in that $d_P$ satisfies all the axioms
of a metric, except that $d_P(A \D B) = 0$ does not necessarily
imply that $A = B$.
In particular, if $A \neq B$ but $A \D B$ has zero measure under $P$,
then $A$ and $B$ are indistinguishable under $P$.
Let us further define a binary relation $\sim_P$ on $\S$ by
\bd
A \sim_P B \iff d_P(A,B) = 0 .
\ed
Then it is easy to verify that $\sim_P$ is an equivalence relation on $\S$.
Also, it is easy to see that an alternate expression for the generalization
error is
\bd
J(T,G_m) = P( A \D B ) = d_P(A,B) .
\ed
Therefore if the hypothesis $G_m$ differs from the target concept $T$
by a set of measure zero (under the chosen probability measure $P$),
then the generalization error would be zero, even though $G_m$
may not equal $T$.
To put it another way, once the probability measure $P$ is specified,
PAC learning tries to identify an element in the equivalence class of $T$
in the quotient space $\C/\sim_P$, and not $T$ itself.

The above discussion explains the limitations of one-bit compressed
sensing as described in \cite{Plan-Versh13b}.
In their case, they choose two vectors in $\R^2$,
namely $x_1 = [ \ba{cc} 1 & 0 \ea ]$ and
$x_1 = [ \ba{cc} 1 & 0.5 \ea ]$.\footnote{They append
a whole lot of zero components which are neglected here.}
Their choice for $P$ is the Bernoulli distribution on $\R^2$, which
is purely atomic and assigns a weight of $0.25$ to the four points
$(1,1), (1,-1),(-1,1),(-1,-1)$.
Now let us plot the half-planes $H_{x_1},H_{x_2}$ and their
symmetric difference, which is the shaded region shown in Figure \ref{fig:0}.
Because none of the four points $(1,1), (1,-1),(-1,1),(-1,-1)$
(shown as red circles)
belongs to the symmetric difference, $x_1$ and $x_2$ are indistinguishable
in OBCS under this probability measure.
In \cite{Plan-Versh13a}, the authors conclude that OBCS cannot always
recover an unknown vector $x$, depending on what $P$ is.
Indeed, $x_1$ and $x_2$ would be indistingishable under \textit{any}
probability measure that assigns a value of zero to $H_{x_1} \D H_{x_2}$.
Therefore, one must be careful to draw the right conclusion:
When subsequent theorems in this paper show
that OBCS is possible under \textit{all} probability measures on $\R^n$,
\textit{including} all purely atomic probability measures, what this
means is that if $x\in \SI_k$, then OBCS will return a vector $\xh$
such that $P(H_x \D H_{\xh}) = 0$ whatever be $P$, and \textit{not}
that $x = \xh$.

\bfig
\bc

\btp[line width=2pt]

\draw [->] (-3,0) -- (3,0) ;
\draw [->] (0,-3) node [below] {$H_{x_1}$} -- (0,3) ;

\draw (-1.5,3) -- (1.5,-3) node [below] {$H_{x_2}$} ;

\draw [line width=1pt,dotted] (-1.5,3) -- (0,3) ;
\draw [line width=1pt,dotted] (-1.0,2) -- (0,2) ;
\draw [line width=1pt,dotted] (-0.5,1) -- (0,1) ;
\draw [line width=1pt,dotted] (0.5,-1) -- (0,-1) ;
\draw [line width=1pt,dotted] (1.0,-2) -- (0,-2) ;
\draw [line width=1pt,dotted] (1.5,-3) -- (0,-3) ;

\filldraw [red] (2,2) circle (2pt) ;
\filldraw [red] (2,-2) circle (2pt) ;
\filldraw [red] (-2,2) circle (2pt) ;
\filldraw [red] (-2,-2) circle (2pt) ;

\etp

\ec
\caption{The half-planes $H_{x_1}$, $H_{x_2}$, and their symmetric difference.}
\label{fig:0}
\efig

Now we examine the relationship of the generalization error
$J(\xh,x)$ to a couple of other quantities that
are widely used in OBCS as error measures.
First, define
\be\label{eq:44}
\r(\xh , x) := E [ | \sg ( \IP{a}{\xh} ) - \sg ( \IP{a}{x} ) | , P ] ,
\ee
where $x$ is the true vector, $\xh$ is its estimate.
Note that $| \sg ( \IP{a}{\xh} ) - \sg ( \IP{a}{x} ) |$ equals $0$ or $2$.
\bd
\r(\xh , x) = 2 J(H_{\xh},H_x) .
\ed

Next, we examine the relationship of $\r(\xh , x)$ to $\nmeu{\xh - x}$.
Without loss of generality it can be assumed that both $x$ and $\xh$ have
unit Euclidean norm.
This can be achieved using some results from \cite{GW95}.
Define
\be\label{eq:44a}
\al := \min_{ \th \in [0,2 \pi]} \frac{2}{\pi} \frac{\th}{1 - \cos \th }
> 0.87856 .
\ee
Then we have the following results.

\begin{lemma}\label{lemma:41}
Let $P$ be any radially invariant probability measure on $\R^n$,
and suppose $a \in \R^n$ is drawn at random according to $P$.
Suppose $\nmeu{x} = \nmeu{\xh} = 1$, and let $J(\xh,x)$ denote
the generalization error defined in \eqref{eq:gen-error}.
Then
\be\label{eq:45}
\nmeusq{\xh - x} \leq \frac{4}{\al} J(\xh,x) .
\ee
\end{lemma}

\textbf{Proof:}
We make use of a couple of results from \cite{GW95}.
First, \cite[Lemma 3.2]{GW95} states that
\bd
\Pr \{ \sg( \IP{a}{\xh}) \neq \sg( \IP{a}{x} ) \} =
\frac{1}{\pi} \arccos (\xh^\top x) ,
\ed
Note that
\bd
J(\xh,x) = \Pr \{ \sg( \IP{a}{\xh}) \neq \sg( \IP{a}{x} ) \} .
\ed
Therefore the above is equivalent to
\bd
J(\xh,x) = \frac{1}{\pi} \arccos (\xh^\top x) .
\ed
Second, \cite[Lemma 3.4]{GW95} states that
\bd
\frac{1}{\pi} \arccos (\xh^\top x)
\geq \frac{\al}{2} ( 1 - \xh^\top x) ,
\ed
or equivalently
\bd
1 - \xh^\top x \leq \frac{2}{\al} \arccos (\xh^\top x) .
\ed
Now note that, when both $\xh$ and $x$ are unit vectors, we have
\bd
\nmeusq{\xh-x} = \nmeusq{\xh} + \nmeusq{x} - 2 \xh^\top x = 2( 1 - \xh^\top x) .
\ed
Therefore
\begin{eqnarray*}
\nmeusq{\xh-x} & = & 2( 1 - \xh^\top x) \\
& \leq & \frac{4}{\al} \arccos (\xh^\top x) 
= \frac{4}{\al} J(\xh,x) .
\end{eqnarray*}
\halmos

\subsection{PAC Learning via the Vapnik-Chervonenkis (VC) Dimension}

One of the most useful concepts in PAC learning theory is defined next.

\begin{definition}\label{def:VC-dim}
A set $S \seq X$ of finite cardinality is said to be {\bf shattered}
by the concept class $\C$ if, for every subset $B \seq S$, there exists
a concept $A \in \C$ such that $S \cap A = B$.
The {\bf Vapnik-Chervonenkis- or VC-dimension} of the concept class $\C$
is the largest integer $d$ such that there exists a set $S$ of cardinality
$d$ that is shattered by $\C$.
\end{definition}

Therefore a concept class $\C$ has VC-dimension $d$ if two statements hold:
(i) There exists a set of cardinality $d$ that is shattered by $\C$,
and \textit{no set} of cardinality larger than $d$ is shattered by $\C$.
If there exist sets of arbitarily large cardinality that are shattered by
$\C$, then its VC-dimension is defined to be infinite.

If a concept class has finite VC-dimension, then
it is PAC learnable under \textit{every probability distribution} on $X$.
An algorithm is said to be \textbf{consistent} if it 
always produces a hypothesis that classifies
all the training samples correctly.
In other words, an algorithm is consistent if the hypothesis $G_m$
produced by applying the algorithm to the sequence
$\{ (c_i , I_T(c_i) ) \}_{i \geq 1}$ has the property that
$I_T(c_i) = I_{G_m}(c_i)$ for all $i$ and $m$.
Note that a consistent algorithm always exists if the axiom of
choice is assumed.
However, in some situations, it is NP-hard or NP-complete to find
a consistent algorithm.

With these notions in place, we have the following
very fundamental result.

\begin{theorem}\label{thm:1}
(\cite{BEHW89}; see also \cite[Theorem 7.6]{MV-03})
Suppose a concept class $\C$ has finite VC-dimension.
Then $\C$ is PAC learnable for every probability measure on $X$.
Suppose that $d$ is an upper bound for $\vcd(\C)$, and let $\{ A_m \}$
be any consistent algorithm.
Then, no matter what the underlying probability measure is,
the learning rate is bounded by
\bd
r(m,\e) \leq 2 \left( \frac{ 2em }{d} \right)^d 2^{-m \e/2} ,
\ed
where $e$ denotes the base of the natural logarithm.
Therefore $r(m,\e) \leq \d$ if
\be\label{eq:samp-comp}
m \geq \max \left\{ \frac{8d}{\e} \lg \frac{8e}{\e} ,
\frac{4}{\e} \lg \frac{2}{\d} \right\}
\ee
samples are chosen.
\end{theorem}

Note that the number of samples required to achieve an accuracy of $\e$
with confidence $1 - \d$ is $O((1/\e) \ln (1/\d))$.
However, the main challenge in applying this result is in finding a
consistent algorithm.

Theorem \ref{thm:1} shows that the finiteness of the VC-dimension of
a concept class is a \textit{sufficient} condition for PAC learnability.
The next result shows that the condition is also necessary.

\begin{theorem}\label{thm:1a}
(\cite{BEHW89,EHKV89}; see also \cite[Theorem 7.7]{MV-03})
Suppose a concept class $\C$ has VC-dimension $d \geq 2$.
Then there exist probability measures on $X$ such that
any algorithm requires at least
\be\label{eq:samp-lower}
m \geq \left\{ \frac{ d-1 }{32 \e} , \frac{1 - \e}{\e} \ln \frac{1}{\d}
\right\}
\ee
samples, in order to learn to accuracy $\e$ and confidence $\d$.
\end{theorem}


\section{Estimates of the VC-Dimension}\label{sec:VC}

The main enabler of the PAC approach to OBCS is an explicit
estimate of the VC-dimension of half-spaces generated by $k$-sparse vectors.

\begin{theorem}\label{thm:2}
Let $\H_k^n$ denote the set of half-spaces $H_k^n(x)$
in $\R^n$ generated by
$k$-sparse vectors, as defined in \eqref{eq:4a}.
Then
\be\label{eq:41}
k ( \lfloor \lg (n/k) \rfloor + 1 ) \leq \vcd ( \H_k^n)
\leq \lfloor 2k \lg ( ne ) \rfloor .
\ee
\end{theorem}

\textbf{Proof:}
We begin with the upper bound in \eqref{eq:41}.
It is shown that if a set $U = \{ u_1 , \ldots , u_l \} \seq \R^n$
is shattered by the collection of half-spaces $\H_k^n$, then
$l \leq \lfloor 2k \lg ( en ) \rfloor$.
The proof combines a few ideas that are standard in PAC learning theory,
which are stated next.

The first result needed is \cite[Theorem 7.2]{Dudley78}.\footnote{Note that
the definition of VC-dimension used in this reference is the VC-dimension
as defined in Definition \ref{def:VC-dim} \textit{plus one}.}
It states the following:
Suppose that $\F$ is a collection of functions mapping a given
set $Z$ into $\R$, such that $\F$ is a $k$-dimensional real vector
space over the field $\R$.
Define the associated collection of subsets of $X$ by
\bd
{\rm Pos}(f) := \{ z \in Z : f(z) \geq 0 \} ,
{\rm Pos}(\F) := \{ {\rm Pos}(f) , f \in \F \} .
\ed
Then $\vcd({\rm Pos}(\F)) = k$.
To apply the above theorem 
to this particular instance, we fix the integer $k$ as well as
a support set $S \seq \{ 1 , \ldots , n \}$ such that $|S| = k$,
and choose $\F$ to be the set of functions
$\{f(z) = \IP{z}{x} \supp(s) \seq S \}$.
This family of functions is clearly a $k$-dimensional linear space
because the adjustable parameter here is the $k$-sparse vector $x$ with
support in the fixed set $S$.
Therefore it follows that, if we define
$\H_S = \{ H_k^n(x) : \supp (x) \seq S \}$,
then $\vcd(\H_S) = k$.

The next result needed is Sauer's lemma \cite{Sauer72}, which states
the following:
Suppose $\C$ is a collection of subsets of $X$ with finite VC-dimension $d$,
and that $U = \{ u_1 , \ldots , u_l \} \seq X$ with $l > d$.
Let $\C \cap U$ denote the collection $\{ A \cap B : A \in \C , B \seq U \}$.
Then
\bd
| \C \cap U | \leq \sum_{i=0}^d \left( \ba{c} l \\ i \ea \right)
\leq \left( \frac{el}{d} \right)^d ,
\ed
where $e$ denotes the base of the natural logarithm.
Strictly speaking, Sauer's lemma is the first inequality, which states
that the number of subsets of $U$ that can be generated by taking
intersections with sets in the collection $\C$ is bounded by the summation
shown.
The second bound is derived in \cite{BEHW89}.
By applying Sauer's lemma to the problem at hand, it can be seen that,
for a fixed support set $S$, the number of subsets of $U$ that
can be generated by intersecting with the collection of half-spaces
$\H_S$ is bounded by $(el/k)^k$, because $\H_S$ has VC-dimension $k$.
Note that a similar bound is derived in \cite{Jacques-et-al13}, but without
any reference to Sauer's lemma.
Also, the result in \cite{Jacques-et-al13} is specifically for collections
of half-spaces, whereas Sauer's lemma is for completely general collections
of sets.

Now observe that $\H_k^n$ is just the union of the collections $\H_S$
as $S$ ranges over all subsets of $\{ 1 , \ldots , n \}$ with $|S| = k$.
The number of such sets $S$ is the combinatorial
parameter $n$ choose $k$, which is bounded by $n^k$.\footnote{Actually
$n^k$ is a pretty crude estimate, but as we shall see, it is
good enough.}
Moreover, for each fixed support set $S$, the collection of subsets
$\H_S \cap U$ has cardinality  no larger than $(en/k)^k$, as shown above.
Therefore
\bd
| \H_k^n \cap U | \leq \left( \ba{c} n \\ k \ea \right)
\left( \frac{el}{k} \right)^k \leq \left( \frac{nel}{k} \right)^k .
\ed

The final step in the proof comes from \cite[Lemma 4.6]{MV-03},
which states the following (see specifically item 2 of this lemma):
Suppose $\al , \beta > 0$, $\al \beta > 4$ and $l \geq 1$.
Then
\be\label{eq:42}
l \leq \al \lg ( \beta l) \implies l < 2 \al \lg ( \al \beta ) .
\ee
In the present instance, the collection of sets $\H_k^n \cap U$
has cardinality no larger $(nel/k)^k$, whereas $U$ has $2^l$ subsets in all.
Therefore, if $U$ is shattered by the collection $\H_k^n$, then we must have
\bd
2^l \leq \left( \frac{nel}{k} \right)^k ,
\ed
or, after taking binary logarithms,
\bd
l \leq k \lg \frac { nel }{k} ,
\ed
which is of the form \eqref{eq:42} with $\al = k, \beta = ne/k$.
Substituting these values into \eqref{eq:42} leads to the conclusion that
\bd
l \leq 2 k \lg ( ne ) ,
\ed
provided $\al \beta = ne \geq 4$, which holds if $n \geq 2$.
Because $l$ is an integer, we can replace the right side by its
integer part, which leads to the upper bound in \eqref{eq:41}.

Now we turn our attention to the lower bound.
First we consider the simple case where $k = 1$.
Given $n$, define $l = \lfloor \lg n \rfloor + 1$, so that $n \geq 2^{l-1}$.
The first step is to show
that the set of half-spaces $\H_1^n$ generated by ``one-sparse
vectors'' has VC-dimension $l$.
Let $s = l-1 = \lfloor \lg n \rfloor$,
and enumerate the $2^s$ bipolar row vectors in $\bp^s$
in some order, call them $v_1 , \ldots v_{2^s}$.
Now define the $n \times l$ matrix
\be\label{eq:43}
M = \left[ \ba{c|c}
1 & v_1 \\ \vdots & \vdots \\ 1 & v_{2^s} \\
0_{( n - 2^s ) \times 1} & 0_{( n - 2^s ) \times s } \ea \right] 
\in \R^{n \times l} .
\ee
In other words, the matrix $M$ has a first column of ones, and then the
$2^s$ bipolar vectors in $\bp^s$ in some order, padded by a block of
zeros in case $n > 2^{l-1}$.
As we shall see below, the ``padding'' is not used.
Now define $U = \{ u_0 , u_1 , \ldots , u_s \}$ denote the $s+1 = l$
columns of the matrix $M$.
Note that for notational convenience we start numbering the columns with 
$0$ rather than $1$.
It is claimed that the collection of half-spaces $\H_1^n$ shatters this set
$U$, thus showing that $\vcd(\H_1^n) \geq l$.

To show that the set $U$ is shattered, let $B \seq U$ be an arbitrary subset.
Thus $B$ consists of some columns of the matrix $M$.
We examine two cases separately.
First, suppose $u_0 \in B$.
Then we associate a unique integer $r$ between $1$ and $2^s$ as follows.
Define a bipolar vector $\i_B \in \bp^s$ by
$i_j = 1$ if $u_j \in B$, and $i_j = -1$ if $u_j \not \in B$.
This bipolar vector $\i_B$ must be one of the vectors $v_1 , \ldots ,
v_{2^s}$.
Let $r$ be the unique integer such that $\i_B = v_r$.
Define the vector $x \in \R^n$ such that $x_r = 1$, and the remaining
elements of $x$ are all zero, and note that $x \in \SI_1$.
Then $\IP{u_0}{x} = 1$, while $\IP{u_j}{x} = 1$ if $u_j \in B$ and
$\IP{u_j}{x} = -1$ if $u_j \not \in B$.
Therefore the associated half-space $H(x)$ includes precisely the elements
of the specified set $B$.
Next, suppose $u_0 \not \in B$; in this case we basically flip the signs.
Thus the bipolar vector $\i_B \in \bp^s$ is chosen such that
$i_j = -1$ if $u_j \in B$, and $i_j = 1$ if $u_j \not \in B$.
If this bipolar vector corresponds to row $r$ in the ordering of $\bp^s$,
we choose $x \in \SI_1$ to have a $-1$ in row $r$ and zeros elsewhere.
This argument shows that the set $\H_1^n$ generated by all one-sparse
vectors $x$ has VC-dimension of at least $\lfloor \lg n \rfloor + 1$,
which is consistent with the left side of \eqref{eq:41} when $k = 1$.

To extend the above argument to general values of $k$, suppose $n$ and $k$
are specified, and define $l = \lfloor \lg (n/k) \rfloor + 1$ and
$s = l - 1 = \lfloor \lg (n/k) \rfloor$.
Then $n/k \geq 2^s$, or equivalently, $n \geq k 2^s$.
Define matrices $M_1 , \ldots , M_k \in \bp^{2^s \times l}$ in analogy
with \eqref{eq:43}.
Then define a matrix
$M \in \bi^{n \times kl}$ as a block-diagonal matrix containing
$M_1 , \ldots , M_k$ on the diagonal blocks, padded by an appropriate
number of zero rows so that the number of rows equals $n$.
In other words, $M$ has the form
\bd
M = \left[ \ba{cccc}
M_1 & 0_{2^s \times l} & \ldots & 0_{2^s \times l} \\
0_{2^s \times l} & M_2 & \ldots & 0_{2^s \times l} \\
\vdots & \vdots & \vdots & \vdots \\
0_{2^s \times l} & \ldots & 0_{2^s \times l} & M_k \\
\multicolumn{4}{c}{0_{ ( n - k 2^s ) \times kl} }
\ea \right] \in \R^{n \times kl} .
\ed
Define $U$ to be the set of columns of the matrix $M$, and note that
$|U| = kl = k( \lfloor \lg (n/k) \rfloor + 1)$.
It is now shown that the set $U$ is shattered by the collection
$\H_k^n$ of half-spaces generated by $k$-sparse vectors.
Partition $U$ as $U_1 \cup U_2 \cup U_k$, where each $U_i$ consists of
$l$ column vectors.
Then any specified subset $B \seq U$ can be expressed as a union
$B_1 \cup \cdots \cup B_k$ where $B_i \seq U_i$ for each $i$.
Now it is possible to mimic the arguments of the previous paragraph
to show that the set $U$ can be shattered by the collection of half-spaces
$\H_k^n$.
For each subset $B_i$, identify an integer $r_i$ between $1$ and $2^s$
such that the bipolar vector $\i_{B_i}$ is the $r_i$-th in the enumeration
of $\bp^s$.
For each index $i$ between $1$ and $k$, let $x_i \in \R^{2^s}$ contain
a $1$ in row $r_i$ and zeros elsewhere.
Define $x \in \R^n$ by stacking $x_1$ through $x_k$, followed by
$n - k 2^s$ zeros.
This shows that it is possible to shatter a set of cardinality
$k ( 1 + \lfloor \lg (n/k) \rfloor )$, which is the right inequality
in \eqref{eq:41}.
\halmos

Theorem \ref{thm:2} is applicable to the case where measurements are
of the form $\sg( \IP{ a_i }{ x } )$.
Such measurements can at best lead to the recovery of the direction of a 
$k$-sparse vector $x$, but not its magnitude.
In situations where it is desired to recover a sparse vector in its entirety,
the measurements are changed to
\be \label{eq:44b}
y_i=\sg( \IP{ a_i }{ x } + b_i \th),
\ee
where $x$ varies over $\SI_k \seq \R^n$ and $\th \in \R$.
The concept class in this case is given by
\be \label{eq:45b}
H_k^{n+1} ( x, \th ) = \{ (a , b) \in \R^{n+1} : \IP{a}{x} 
+ b \th \geq 0 \} .
\ee
By inspecting the equation \eqref{eq:44b}, we deduce that
\be \label{eq:46}
\H_k^{n} ( x ) \subseteq \H_k^{n+1} ( x ) \subseteq \H_{k+1}^{n+1} ( [x ~ \th]^\top ),
\ee
where $x \in \Sigma_k$ and $\th \in \R$ so that the vector
$ [x ~ \th]^\top$ is $k+1$-sparse.
We are now ready to state the first result of this section.

\begin{theorem}\label{thm:4}
Let $\H_k^{n+1}$ denote the set of half-spaces $H_k^n(x)$
in $\R^n$ as defined in \eqref{eq:44b}.
Then
\be
k ( 1 + \lfloor \lg (n/k) \rfloor ) \leq \vcd ( \H_k^n)
\leq \lfloor 2(k+1) \lg ( e(n+1) ) \rfloor .
\ee
\end{theorem}

\textbf{Proof:}
From \eqref{eq:46} we conclude that
\[
 \vcd H_k^{n} ( x )  \leq \vcd ( \H_k^{n+1}) \leq \vcd H_{k+1}^{n+1} ( [x ~ z]^t ),
\]
then the desired result follows Theorem~\ref{thm:2}.
\halmos

\section{OBCS with Noisy Measurements}\label{sec:noisy}

In this section we study the one-bit compressed sensing problem
when the information available to the learner is a noisy version
of the true output $\sg( \IP{a_i}{x})$ or $\sg( \IP{a_i}{x} + b_i)$,
where $x$ is an unknown $k$-sparse vector.
Specifically, the label $y_i$ equals this sign with probability
$1 - \al$, and gets flipped with probability $\al$, where $\al \in (0,0.5)$.
In the PAC learning literature, the problem of concept learning with
mislabelling has been studied, and there are some papers on the topic.
Rather than cite these, we refer the reader to a recent paper
\cite{Natarajan-et-al13} and the references therein.
In this paper, which represents the state of the art,
it is assumed that the error rate $\al$ is known.
By adopting a different approach, we are able to show that minimizing
empirical risk leads to provably near-optimal estimates, even without
knowing $\al$.
Therefore the results given here are of independent interest.
Note that in \cite{Natarajan-et-al13}, it is \textit{not} assumed
that the two error probabilities (namely a one becoming a zero  and vice versa)
are equal.
This assumption is made here \textit{solely in the interests of convenience},
and can be dispensed with at the expense of more elaborate notation.

To make the problem formulation precise, we use the notation in
Section \ref{ssec:PAC}, whereby $X$ is a set, $\S$ is a $\s$-algebra
of subsets of $X$, and $P$ is a probability measure on $X$.
To incorporate the randomness, we enlarge $X$ by defining $X_N = X \times \bi$
as the sample space; define $\S_N$ to be the $\s$-algebra of subsets
in $X_N$ generated by cylinder sets of the form $S \times \{ 0 \}$
and $S \times \{ 1 \}$ for all $S \in \S$; and define a probability
measure $P_N$ on $X_N$ by defining
\be\label{eq:51}
P_N( S \times \{ 0 \} ) = (1 - \al ) P(S) ,
P_N( S \times \{ 1 \} ) = \al P(S) .
\ee
Let $(c,L)$ denote a typical element in the sample space $X_N$.
Then
\bd
\Pr \{ L = 0 \} = P_N( X \times \{ 0 \} ) = 1 - \al ,
\Pr \{ L = 1 \} = P_N( X \times \{ 1 \} ) = \al .
\ed
Here the event $L = 0$ corresponds to the label not being flipped,
while the event $L = 1$ corresponds to the label being flipped.
It is clear that $P_N$ is a product measure, so that the flipping of
labels is independent of the generation of training samples.

Learning takes place as follows:
Independent samples $\{ c_i,\om_i \}_{i \geq 1}$ are generated
in accordance with the above probability measure $P_N$.
Let $T$ be a fixed but unknown target concept.
Then for each $i$, a label $y_i$ is generated as
\be\label{eq:52}
y_i = | I_T(c_i) - \om_i | .
\ee
This is equivalent to saying that $y_i = I_T(c_i)$ with probability
$1 - \al$ and $y_i = 1 - I_T(c_i)$ with probability $\al$.
As before, an algorithm is an indexed family of maps 
$A_m : ( X \times \bi )^m \ap \C$ for each $m \geq 1$.
The algorithm $A_m$ is applied to the set of labelled samples
$\{ (c_i,y_i) \}_{i=1}^m$, giving rise to a hypothesis $G_m$.

To assess how well a hypothesis $F$ (however it is derived)
approximates the unknown target concept $T$,
we generate a random test input $x \in X$ according to $P$, and then
predict that the oracle output on $x$ will be $I_F(x)$.
The error criterion therefore equals
\be\label{eq:53}
J_N(T,F) := E[ | f(I_T(x)) - I_F(x) | , P_N ] ,
\ee
where $f(I_T(x))$ is the noisy label and $I_F(x)$ is the indicator
function of $F$.
The premise in the above definition is that, while the oracle output
is noisy, our prediction is not noisy.

The main difference from the case of noise-free labelling is that,
even if $F$ were to equal $T$, the error $J_N(T,F)$
would not equal zero, due to the noisy labelling.
Note that, for a given $x \in X$, the quantity $| f(I_T(x)) - I_F(x) |$
equals $| I_T(x) - I_F(x) |$ with probability $1 - \al$,
and equals $1 - | I_T(x) - I_F(x) |$ with probabilty $\al$.
Therefore
\beq
J_N(T,F) & = & \int_X [ (1 - \al) | I_T(x) - I_F(x) |
+ \al ( 1 - | I_T(x) - I_F(x) | ) ] P(dx) \nonumber \\
& = & \al + (1 - 2 \al) \int_X | I_T(x) - I_F(x) | P(dx) \nonumber \\
& = & \al + (1 - 2 \al) d_P(T,F) . \label{eq:53a}
\eeq
In the case of noise-free labelling,
the minimum achievable value of the error measure $J$ (as defined in
\eqref{eq:gen-error}) is $0$, which is
achieved by any hypothesis $F$ such that $P(T \D F) = 0$.
In contrast, the minimum achievable value of the modified error measure
$J_N$ is $\al$, which is again
achieved by any hypothesis $F$ such that $P(T \D F) = 0$.
Therefore, to measure the performance of an algorithm with noise-corrupted
labels, one should compare the error $J_N$ with the minimum achievable
value of $\al$.
This motivates the next definition.
Let $G_m$ denote the hypothesis generated by the algorithm, and set
\be\label{eq:rate-e}
r_N(m,\e) := \sup_{T \in \C} P^m \{ \c \in X^m : J_N(T,F_m) > \al + \e \} .
\ee
Note that if $\al = 0$ so that the measurements are noise-free,
then $J_N$ reduces to $J$ as defined in \eqref{eq:gen-error}
and $r_N(m,\e)$ reduces to $r(m,\e)$ as defined in \eqref{eq:rate}.

\begin{definition}\label{def:PAC-N}
An algorithm $\{ A_m \}$ is said to be {\bf probably approximately
correct (PAC) with noise-corrupted measurements}
if $r_N(m,\e) \ap 0$ as $\mai$, for every fixed $\e > 0$.
The concept class $\C$ is said to be {\bf PAC learnable with noise-corrupted
measurements} if there exists a PAC algorithm.
\end{definition}

In the case of noise-free measurements, the results on learnability
were stated in terms of a consistent algorithm, which always exists
if one were to assume the axiom of choice.
In contrast, in the case where the labels are noisy, it might not be
possible to construct a hypothesis that is consistent.
Therefore the notion of consistency is replaced by the notion of
minimizing empirical risk.
Suppose we are given a labelled sample sequence $\{ (c_i, y_i) \in
X \times \bi \}_{i \geq 1}$.  Suppose $F \in \C$ is a hypothesis.
Then the \textbf{empirical risk} of the hypothesis with respect to this
labelled sequence, after $m$ samples, is defined as
\be\label{eq:54}
\Jh_m(T,F) := \frac{1}{m} \sum_{i=1}^m | y_i - I_F(c_i) | .
\ee

\begin{definition}\label{def:MER}
An algorithm $\{A_m\}_{m \geq 1}$is said to
\textbf{minimize empirical risk}, or to be a
\textbf{MER algorithm}, if for all sample sequences $\{ (c_i, y_i) \in
X \times \bi \}_{i \geq 1}$, and all integers $m$, it is the case that
\be\label{eq:55}
\Jh_m(G_m) = \min_{F \in \C} \Jh_m(F) ,
\ee
where
\bd
G_m = A_m( (c_1 , y_1) , \ldots (c_m,y_m))
\ed
is the output of the algorithm after $m$ samples,
given the sequence $\{ (c_i, y_i) \in X \times \bi \}_{i \geq 1}$.
\end{definition}

Note that if the labels are noise-free, then $y_i = I_T(c_i)$, and
\bd
\Jh_m(T,F) := \frac{1}{m} \sum_{i=1}^m | I_T(c_i) - I_F(c_i) | 
\ed
is the empirical estimate of the distance $D_P(T,F)$.
In this case, a MER algorithm becomes a consistent algorithm.

Now we state the main result regarding PAC learning with noisy labels.

\begin{theorem}\label{thm:51}
Suppose $P \in \P^*$ is an arbitrary probability measure on $X$,
and suppose that $\vcd(\C) \leq d$.
Let $T \in \C$ be any fixed but unknown target concept,
and let $G_m$ be the output of a MER algorithm after $m$ labelled samples.
Then
\be\label{eq:56}
\Pr \{ J(T,G_m) > \al + \e \} \leq c(m,\e),
\ee
where
\be\label{eq:57}
c(m,\e) = \left[ 4 \left( \frac{ 0.2 e m }{d} \right)^{10d} + 1 \right]
\exp(-0.08 m \e^2) .
\ee
\end{theorem}

\textbf{Remarks:} Note that the mislabelling probability $\al$ 
\textit{does not appear} in the bound $c(m,\e)$.
Therefore, unlike other results in this area, our bound does not
require one to know $\al$, and is thus universal.
Note too that some authors would write $J^*$ to denote the minimum
achievable loss function, just to make it explicit that the bound
is on the probability that the hypothesis is suboptimal by a quantity $\e$.

The proof of Theorem \ref{thm:51} proceeds through a series of
preliminary results.
The basic approach is to expand the original probability space $(X,\S,P)$
to $(X_N,\S_N,P_N)$ where $X_N = X \times \bi$ etc., as described above.
The next step is to note that the quantity $\Jh_m(T,F)$ is just the
empirical estimate of $J(T,F)$, under the extended probability measure $P_N$.
It is shown that these empirical estimates converge
uniformly to their true values, as a consequence of the assumption that
the concept class $\C$ has finite VC-dimension.
This is a rough outline of the approach to be pursued.

We begin by identifying all subsets of $X_N$ that could cause the label
$y_i$ to equal one, as $T$ varies over $\C$.
Note that if $c_i \in T$ and $\om_i = 0$, then $y_i$ would equal one.
Alternatively, if $c_i \not \in T$ and $\om_i = 1$, then too
$y_i$ would equal one.
Therefore we identify two collections of subsets of $X_N$, which we
denote by $\C_N$ and $\C_N^c$ respectively, namely
\bd
\C_0 = \{ T \times \{ 0 \} : T \in \C \} ,
\C_1^c = \{ T^c \times \{ 1 \} : T \in \C \} .
\ed

Next, observe that $I_F$ equals one if and only if $c_i \in F$, or
equivalently, $(c_i,\om_i) \in F \times \bi$.
So we can define
\bd
\C_N = \{ F \times \bi : F \in \C \}
\ed
to be the collection of all subsets of $X_N$ that could cause the label
$I_F$ to equal one.

\begin{lemma}\label{lemma:51}
Suppose $\vcd(\C) \leq d$.
Then $\vcd( \C_0 \cup \C_1^c ) = \vcd(\C) \leq d$, 
and $\vcd(\C_N) = \vcd(\C) \leq d$.
\end{lemma}

\textbf{Proof:}
Suppose $S_N = \{ (c_1 , \om_1) , \ldots , (c_s,\om_s) \} \seq X_N$
is shattered by $\C_0 \cup \C_1^c$.
If $A_N \in \C_0$, then the second component of all elements of $A_N$ is zero.
Similarly, if $A_N \in \C_1^c$, then the second component of all elements
of $A_N$ is one.  
Therefore, if both $0$ and $1$ occur in the list $\{ \om_1 , \ldots , \om_s)$,
then such a set $_N$ cannot be shattered by $\C_0 \cup \C_1^c$.
To see why, renumber the components such that $\om_1 = 0$ and $\om_2 = 1$.
Then there cannot exist a set $A_N$ in either $\C_0$ or $\C_1^c$ such that
$A_N \cap S_N = \{ (c_1 , 0) , (c_2 , 1 ) \}$.
Therefore, if $S_N \seq X_N$ is shattered by $\C_0 \cup \C_1^c$,
it must be of the form $S \times \{ 0 \}$ or $S \times \{ 1 \}$,
where $S = \{ c_1 , \ldots , c_s \} \seq X$.
If $S_N$ is of the form $S \times \{ 0 \}$, then it can only be
shattered by $\C_0$, which implies that $S$ itself is shattered by $\C$.
Therefore $|S| \leq d$.
By entirely analogous reasoning, if 
$S_N$ is of the form $S \times \{ 1 \}$, then it can only be
shattered by $\C_1^c$, which implies that $S$ itself is shattered by $\C^c$.
Now we make use of the easily proved fact that $\C$ and $\C^c$ shatter
exactly the same sets.
Therefore once again $|S| \leq d$.
In either case $|S_N| \leq d$.
This leads to the first conclusion.
The proof that $\vcd(\C_N) = \vcd(\C)$ is similar and is omitted.
\halmos


\begin{theorem}\label{thm:52}
Suppose $\vcd(\C) \leq d$, and let $T , F \in \C$ be arbitrary.
Also let $P \in \P^*$ be an arbitrary probability measure.
Then
\be\label{eq:58}
\sup_{T,F \in \C} \Pr \{ | J(T,F) - \Jh_m(T,F) | > \e \}
\leq 4 \left( \frac{ 0.2 e m }{d} \right)^{10d} \exp(-m \e^2 / 8) .
\ee
\end{theorem}

\textbf{Proof:}
Let, as before, $\C_0 \cup \C_1^c$ denote the collection of sets
that generate a label $y_i$ of one.
Similarly, $\C_N$ is the collection of sets that generate a label $I_F$ of one.
Given concept classes $\A_N , \B_N\seq \S_N$, define
\bd
\A_N \D \B_N := \{ A_N \D B_N : A_N \in \A_N , B_N \in \B_N \} .
\ed
Then it follows from \cite[Thorem 4.5]{MV-03} applied with $k = 2$, that
\bd
\vcd(\A_N \D \B_N) \leq 10 \max \{ \vcd(\A_N) , \vcd(\B_N) \} .
\ed
Then $\Jh$ as defined in \eqref{eq:54} is the empirical distance
between two sets, one belonging to $\C_0 \cup \C_1^c$ and the
other belonging to $\C_N$.
Now $\vcd(\C) \leq d$, and in turn this implies that
$\vcd(\C_0 \cup \C_1^c ) \leq d$ by Lemma \ref{lemma:51}.
Therefore
\bd
\vcd[ ( \C_0 \cup \C_1^c ) \D \C_N ] \leq 10 d.
\ed
Now observe that $J(T,F)$ is the expected value of $|y - I_F|$,
while $\Jh_M$ is an empirical mean based on $m$ samples.
Therefore \eqref{eq:58} follows from \cite[Theorem 7.4]{MV-03}.
\halmos

Now we give the proof of Theorem \ref{thm:51}.

\textbf{Proof:}
We begin by establishing an elementary result.
Suppose $X_1, X_2$ are random variables, not necessarily independent,
and that $\e_1, \e_2$ are thresholds.
Then
\be\label{eq:59}
\Pr \{ X_1 + X_2 > \e_1 + \e_2 \} \leq
\Pr \{ X_1 > \e_1 \} + \Pr \{ X_2 > \e_2 \} .
\ee
To see this, note that
\bd
X_1 \leq \e_1 \mbox{ and } X_2 \leq \e_2 \imp X_1 + X_2 \leq \e_1 + \e_2 .
\ed
Therefore
\bd
\Pr \{ X_1 + X_2 \leq \e_1 + \e_2 \} \geq 
\Pr \{ X_1 \leq \e_1 \mbox{ and } X_2 \leq \e_2 \} .
\ed
Taking the contrapositive shows that
\begin{eqnarray*}
\Pr \{ X_1 + X_2 > \e_1 + \e_2 \} & \leq &
\Pr \{ \neg ( X_1 \leq \e_1 \mbox{ and } X_2 \leq \e_2 ) \} \\
& = & \Pr \{ X_1 > \e_1 \mbox{ or } X_2 > \e_2 \} \\
& \leq & \Pr \{ X_1 > \e_1 \} + \Pr \{ X_2 > \e_2 \} .
\end{eqnarray*}

Returning to the theorem, suppose $T$ is the target concept and
$G_m$ is the hypothesis produced by a MER algorthm.
Now, because $G_m$ is the output of a MER algorthm, we have that
\be\label{eq:510}
\Jh(G_m,T) \leq \Jh(T,T) ,
\ee
where $\Jh(T,T)$ denotes the empirical distance 
\bd
\Jh(T,T) = \frac{1}{m} | y_i - I_T(c_i) | .
\ed
Note the right side is a random variable with an expected value of $\al$
(the probability of the label $I_T(c_i)$ being flipped).
Therefore, by the additive form of the Chernoff bound (see for example
\cite[p.\ 24]{MV-03}), it follows that
\be\label{eq:511}
\Pr \{ \Jh(T,T) > \al + \e_1 \} \leq \exp(-2 m \e_1^2 ) , \fa \e_1 .
\ee
Combining \eqref{eq:510} and \eqref{eq:511} shows that
\be\label{eq:512}
\Pr \{ \Jh(G_m,T) > \al + \e_1 \} \leq \exp(-2 m \e_1^2 ) , \fa \e_1 .
\ee
Next, due to the uniform convergence property of empirical means to
their true values, it follows that
\be\label{eq:513}
\Pr \{ J(T,G_m) - \Jh(T,G_m) > \e_2 \}
\leq 4 \left( \frac{ 0.2 e m }{d} \right)^d \exp(-m \e_2^2 / 8) , \fa -e_2 .
\ee
Now, given a threshold $\e$, we can choose any $\e_1 , \e_2$ such that
$\e = \e_1 + \e_2$, and apply the above bounds.
We choose $\e_1 = 0.2 \e , \e_2 = 0.8 \e$, so that the two exponents match.
This leads to
\begin{eqnarray*}
\Pr \{ J(T,G_m) > \al + \e \} & \leq &
\Pr \{ \Jh(G_m,T) > \al + \e_1 \} +
\Pr \{ J(T,G_m) - \Jh(T,G_m) > \e_2 \} \\
& = & c(m,\e) .
\end{eqnarray*}
This completes the proof.
\halmos

\section{Algorithm for One-Bit Compressed Sensing}\label{sec:alg}

By combining Theorem \ref{thm:1a} with Theorem \ref{thm:2} on
the \textit{lower} bound on the VC-dimension of half-spaces generated
by $k$-sparse $n$-vectors, we can prove the following result:

\begin{theorem}\label{thm:2b}
There exists a probabiity measure $P$ on $\R^n$ such that
\textit{any} algorithm that leads to a uniform error estimate of the
form $J(\xh,x) \leq \e$ for all $k$-sparse vectors $x \in \R^n$
requires at least $\OM(k \lg(n/k))$ samples.
\end{theorem}

While this theorem might be only of theoretical interest, it does show
the intrinsic difficulty of OBCS, which no algorithm can cross.

Now let us study how to solve the OBCS problem.
The results in Theorems \ref{thm:1} and \ref{thm:2} can be combined
to produce the following ``conceptual'' algorithm.

\begin{theorem}\label{thm:2a}
Let integers $n,k$ with $k \ll n$ be specified, and suppose $x \in \R^n$
is $k$-sparse.
Let $P$ be an arbitrary probability distribution on $R^n$, and choose
$\{ a_i \}_{i \geq 1}$ generated independently at random according to $P$.
Let $y_i = \sg(\IP{a_i}{x})$ for all $i$.
With these conventions, any algorithm that generates a $k$-sparse estimate $\xh$
such that $y_i = \sg( \IP{a_i}{\xh} )$ for all $i$ is probably approximately
correct.
In particular, given an accuracy $\e$ and a confidence $\d$, let
\bd
\delta = \lfloor 2k \lg ( ne ) \rfloor ,
\ed
and choose at least $m$ samples where
\be\label{eq:61}
m \geq \max \left\{ \frac{8d}{\e} \lg \frac{8e}{\e} ,
\frac{4}{\e} \lg \frac{2}{\d} \right\} ,
\ee
Then it can be guaranteed with confidence of at least $1 - \d$ that
$J(\xh,x) \leq \e$, and $\r(\xh,x) \leq 2 \e$.
Moreover, if $P$ is a radially invariant probability distribution on $\R^n$,
Then it can be guaranteed with confidence of at least $1 - \d$ that
\be\label{eq:62}
\nmeusq{\xh - x} \leq \frac{4}{\al} \e ,
\ee
where the constant $\l$ is defined in \eqref{eq:44a}.
\end{theorem}

In the case where the labels $y_i$ are noisy versions of $I_T(c_i)$,
the above theorem can be modified to say that any MER algorithm is
PAC, using Theorem \ref{thm:51}.

Let us return to the conceptual algorithm outlined above.
Suppose we are given randomly generated labelled samples
$\{ ( a_i , y_i ) \}_{i=1}^m$, where $a_i \in \R^n$ and $y_i \in \bp$,
where $y_i$ is a possibly noise-corrupted measurement of $\sg(\IP{a_i}{x})$.
Define
\bd
\M = \{ 1 , \ldots , m \} , \M_+ = \{ i \in \M : y_i = 1 \} ,
\M_- = \{ i \in \M : y_i = -1 \} .
\ed
Ideally, in the case where measurements are noise-free,
we would like to find a $k$-sparse vector $\xh$ such that
\bd
\IP{a_i}{\xh} > 0 \fa i \in M_+ , \IP{a_i}{\xh} < 0 \fa i \in \M_- .
\ed
This would lead to a ``consistent'' hypothesis.
If there exists a vector $\xh \in \R^n$ that satisfies the above inequalities,
the data is said to be linearly separable.
However, finding a \textit{$k$-sparse} separating vector $\xh$ may not be
easy.
In the case of noisy measurements, 
constructing a MER algorithm would require us to find
an $\xh$ ($k$-sparse or otherwise) such that
the number of violations in the above inequality is minimized.
However, this problem is NP-hard, as shown in \cite{Natarajan95}.
Therefore we need to look for alternate approaches that do not strictly
conform to the theory.

It is proposed here to use the $\ell_1$-norm support vector machine
(SVM) formulation as introduced in \cite{Bradley-Mangasarian98},
which is a modification of the
the widely used $\ell_2$-norm (SVM) formalism
introduced in \cite{Cortes-Vapnik97}.
This formulation of the $\ell_1$-norm SVM has two important advantages
over the standard $\ell_2$-norm SVM formlation in \cite{Cortes-Vapnik97}.
First, the weight vector generated by the $\ell_1$-norm SVM is sparse,
unlike with the $\ell_2$-norm SVM.
Second, the particular formulation suggested in 
\cite{Bradley-Mangasarian98} \textit{works even when the data is
not linearly separable}.
So we begin by describing the $\ell_1$-norm SVM, before proceeding to our
proposed algorithm.

Then the modified $\ell_1$-norm SVM formulation in
\cite{Bradley-Mangasarian98} can be stated as follows:
\bd
\xh = \argmin_{z \in \R^n}
(1 - \l) \left[ \sum_{i=1}^{m_1} \al_i + \sum_{i=1}^{m_2} \beta_i \right]
+ \lambda \sum_{i=1}^n | z_i | \st
\ed
\bd
\IP{a_i}{z} + \al_i \geq 1 \fa i \in \M_+ ,
\IP{a_i}{z} - \beta_i \leq -1 \fa i \in \M_j ,
\ed
\be\label{eq:63}
\al_i \geq 0 \fa i \in \M_+ , \beta_i \geq 0 \fa i \in \M_- .
\ee
There are a few points to note here.
First, the constraints have ``slack'' variables $\al_i, \beta_i$
so that problem formulation makes sense even when the data is not
linearly separable.
This is in contrast to the standard SVM formulation
\be\label{eq:64}
\min_{z \in \R^n} \sum_{i=1}^n | z_i | \st
\IP{a_i}{z} \geq 1 \fa i \in \M_+ ,
\IP{a_i}{z} \leq -1 \fa i \in \M_- .
\ee
Note that the formulation in \eqref{eq:64} is equivalent to that in
\cite{Plan-Versh13a}, because they use the normalization
$\sum_{i=1}^m | \IP{a_i}{z} | = m$, while in \eqref{eq:64} the normalization
is that the minimum ``gap'' in satisfying the constraints equals $1$.
If the measurements are of the type $y_i = \sg(\IP{a_i}{x} + b_i)$,
Then the problem formulation is modified to
\bd
(\xh , \hat{v}) = \argmin_{(z,v)} 
(1 - \l) \left[ \sum_{i=1}^{m_1} \al_i + \sum_{i=1}^{m_2} \beta_i \right]
+ \lambda \sum_{i=1}^n | z_i | + \t |v| \st
\ed
\bd
\IP{a_i}{z} + \al_i \geq 1 \fa i \in \M_+ ,
\IP{a_i}{z} - \beta_i \leq -1 \fa i \in \M_j ,
\ed
\be\label{eq:65}
\al_i \geq 0 \fa i \in \M_+ , \beta_i \geq 0 \fa i \in \M_- .
\ee
This is different from the formulation suggested in
\cite{KSW16}, as described here in \eqref{eq:22}, in exactly the same
manner that the formulation in \cite{Plan-Versh13a}
differs from \eqref{eq:63}.
The major advantage of \eqref{eq:63} over the formulation in
\cite{Plan-Versh13a}, or of \eqref{eq:65} over the formulation in
\cite{KSW16} is this:
In case the labels $y_i$ are corrupted by measurement noise,
both the formulations of \cite{Plan-Versh13a} and \cite{KSW16}
would be infeasible, whereas the formulations in \eqref{eq:63} 
and \eqref{eq:65} continue to be meaningful.
The second point follows upon the first.
The objective function is the sum of the slack variables
and the $\ell_1$-norm of $z$.
Including the $\ell_1$-norm of $z$ in the objective function 
forces the solution $z$ to be sparse.
The constant $\lambda \in (0,1)$ provides a trade-off between minimizing
$\nmm{z}_1$ and violating the constraints.
By choosing the weight $\lambda \in (0,1)$ is close to, but not equal to, zero,
one can ensure that the optimization attempts
to violate the constraints by as little as possible.

The above formulation does not lead to an $\xh$ that is $k$-sparse.
The next step is to \textit{truncate} $\xh$ by retaining only the $k$
largest components by magnitude and discarding the next.

\section{A Numerical Example}\label{sec:exam}

We chose $n = 1000, k = 20$, and generated 30 vectors $x \in \SI_k$ at random.
Then we generated $m$ measurements of the form $\sg(\IP{a_i}{x} + b_i)$
where $a_i, b_i$ are normally distributed.
Then we applied the algorithm proposed in the previous section, that is,
carrying out the minimization in \eqref{eq:65} and then truncating
the resulting $\xh$ to the $k$ dominant components by magnitude.

Figure \ref{fig:1} shows a comparison of the error for our method versus
that for the method proposed in \cite{KSW16}, for one of the randomly
generated $k$-sparse vectors.
It is evident that our method ourperforms the latter.
This can perhaps be attributed to replacing the conventional SVM
as in \cite{KSW16} with the modification in \eqref{eq:65}.
Figure \ref{fig:2} shows the number of correctly recovered components
as a function of $m$, for one of the randomly generated $k$-sparse vectors.
Figure \ref{fig:3} shows the box plot of the mean square error as
a function of $m$ over all $30$ random repetitions of this experiment.
The small horizontal lines show the maximum and minimum error, while
the boxes display the 25th and 75th percentiles.
Note that, in OBCS, it is meaningful to have $m > n$.

\bfig
\bc
\includegraphics[width=120mm]{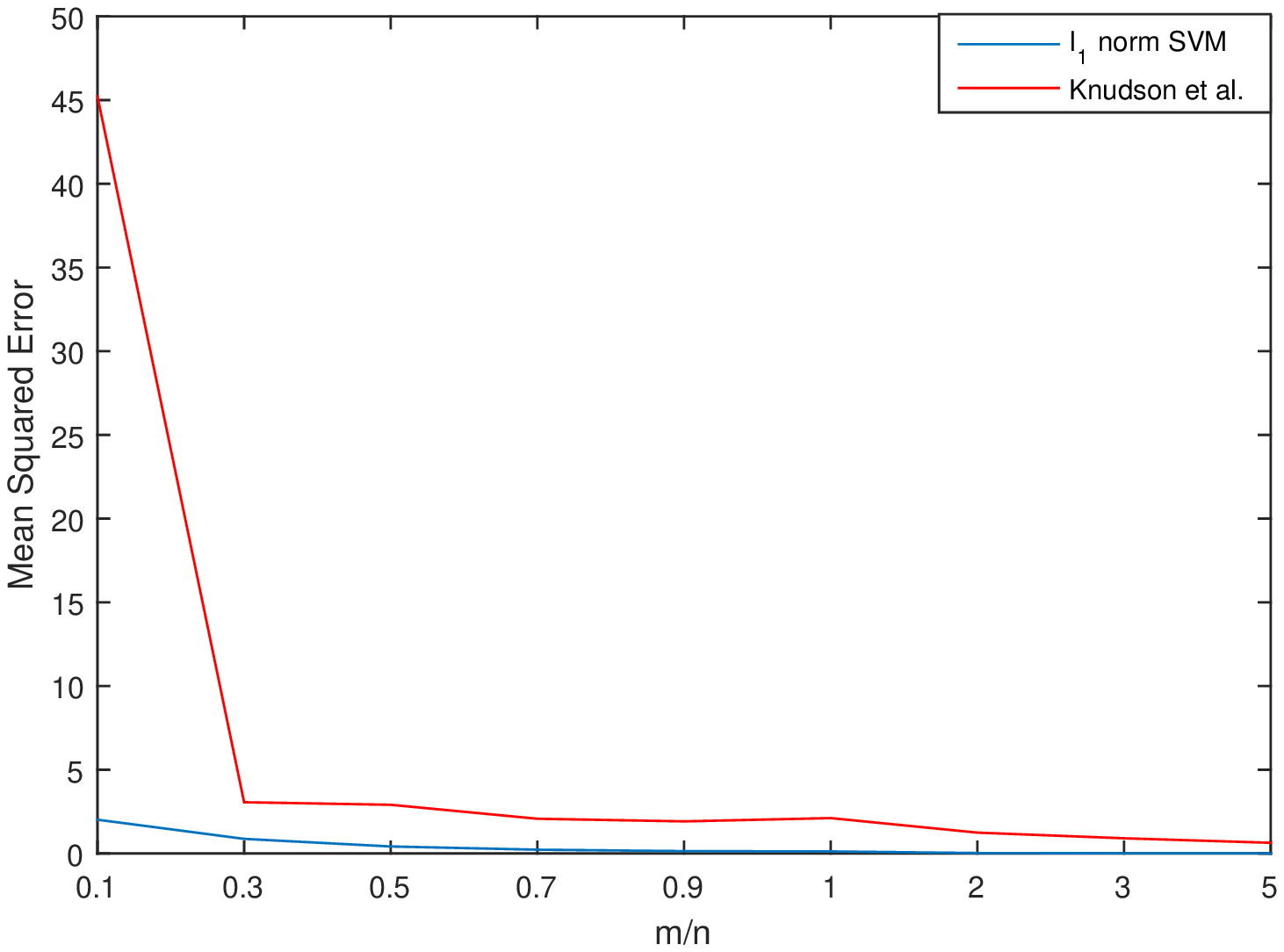}
\ec
\caption{Mean square error between $\xh$ and $x$ as a function of $m$
for our algorithm and that of \cite{KSW16}}
\label{fig:1}
\efig

\bfig
\bc
\includegraphics[width=120mm]{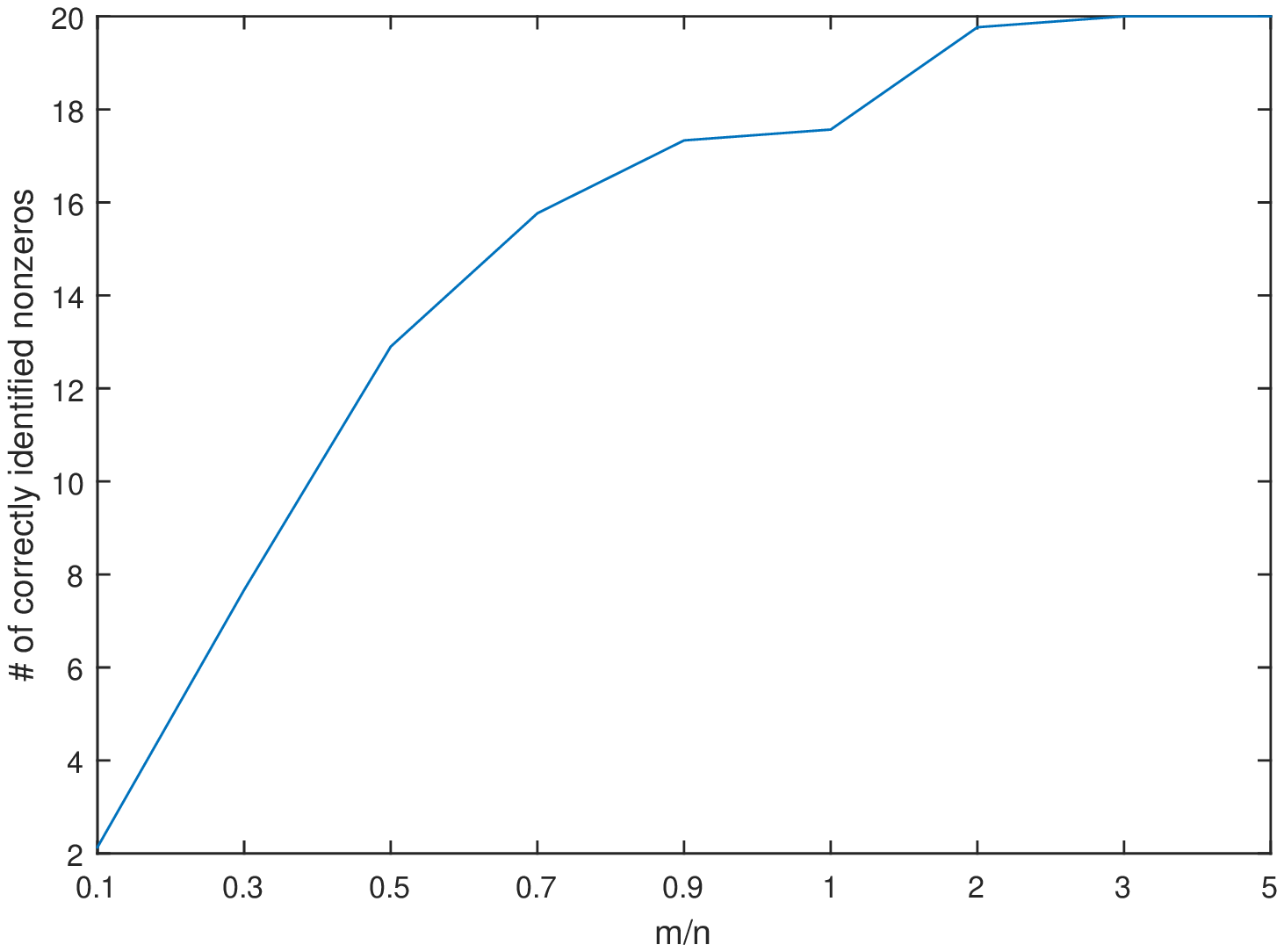}
\ec
\caption{Number of correctly recovered  components (out of 20)
as a function of $m$}
\label{fig:2}
\efig

\bfig
\bc
\includegraphics[width=120mm]{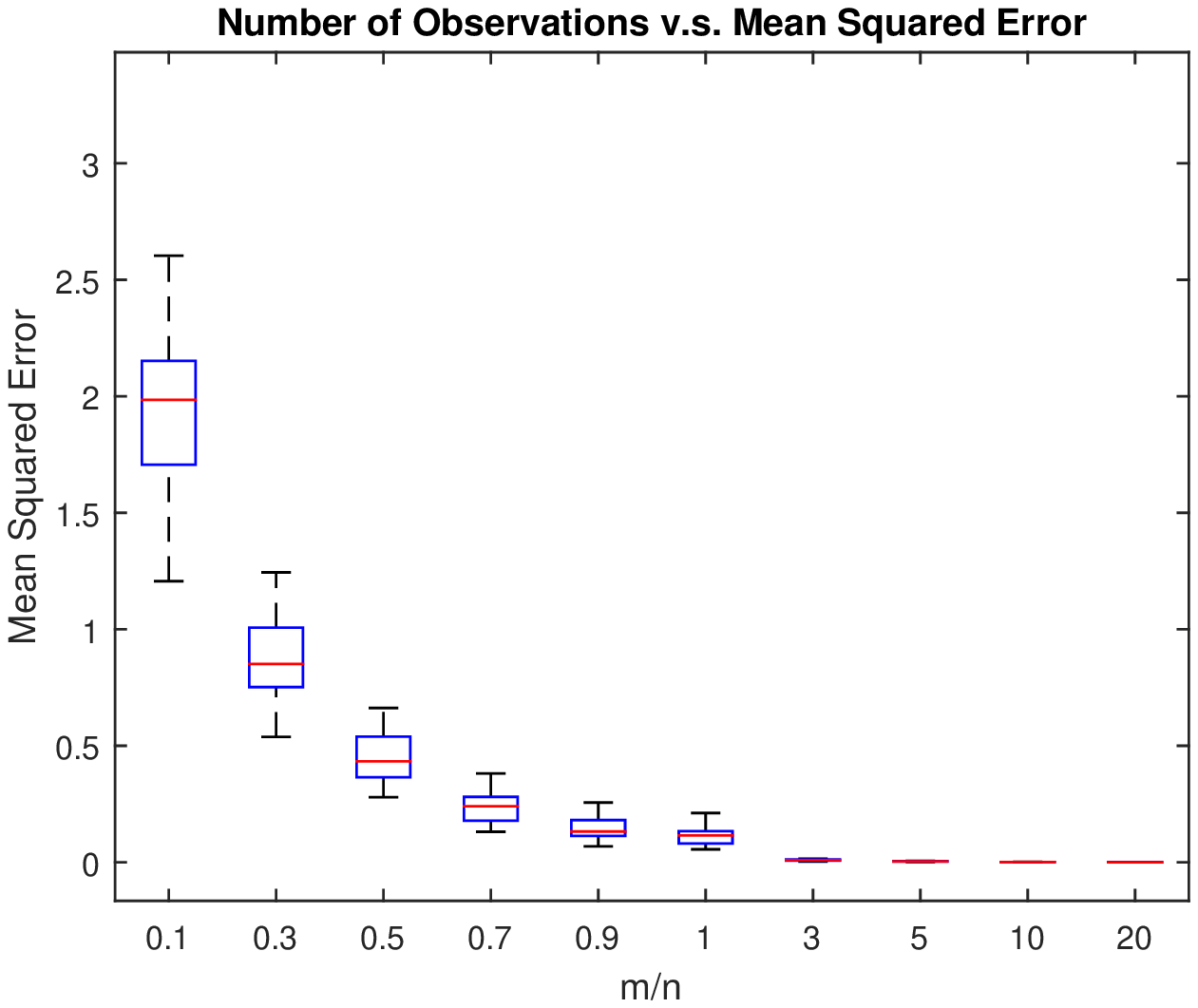}
\ec
\caption{Mean square error as a function of $m$: Box plot}
\label{fig:3}
\efig

\section{Discussion}\label{sec:disc}

In this paper, the problem of one-bit compressed sensing (OBCS) has been
formulated as a problem in probably approximately correct (PAC) learning
theory.
In particular, it has been shown that the VC-dimension of the set of
half-spaces in $\R^n$ generated by $k$-sparse vectors is bounded by
$O(k \lg (n/k))$.
Therefore, in principle at least, the OBCS problem can be solved
using only $O(k \lg (n/k))$ samples.
This is possible in principle even when the measurements are corrupted
by noise.
However, in general, it is NP-hard to find a consistent algorithm
when measurements are free from noise, and to find an algorithm
that minimizes empirical risk when measurements are noisy.
We proposed a modification of the $\ell_1$-norm support vector machine
as a feasible alternative, and illustrated that our approach outperforms
earlier algorithms in the literature.

One of the main advantages of formulating OBCS as a problem in PAC
learning is that extending these results to the case where the 
samples $\{ a_i \}$ (or $\{ (a_i,b_i) \}$ as the case may be)
are not i.i.d.\ essentially ``comes for free.''
It is now known that, if a concept class has finite VC-dimension,
then empirical means converge to their true values not only for i.i.d.\
samples $\{ a_i \}$ (or $\{ (a_i,b_i) \}$ as the case may be),
but also when this sequence forms an ergodic process; 
see \cite{Adams-Nobel10}, which builds on an earlier result in
\cite{Nobel-Dembo93} for $\beta$-mixing processes.
However, in order to be useful in OBCS, it is not enough to know this.
One must also have expicit estimates of the \textit{rate} at which
empirical means converge to their true values, or what is called
the learning rate here.
Such estimates are provided in \cite{RLK-MV02}.
As it is fairly straight-forward to adapt the various theorems given
here to the case of $\beta$-mixing processes using the above-mentioned
results, the details are omitted.


\bibliographystyle{IEEEtran}

\bibliography{Comp-Sens}

\end{document}